\documentclass{article}

\PassOptionsToPackage{numbers}{natbib}
\usepackage[preprint]{template}
\usepackage[utf8]{inputenc}
\usepackage[T1]{fontenc}
\usepackage{url}
\usepackage{graphicx}
\usepackage{booktabs}
\usepackage{array}
\usepackage{amsmath}
\usepackage{amssymb}
\usepackage{mathtools}
\usepackage{microtype}
\usepackage{xcolor}
\usepackage{float}
\usepackage{tikz}
\usetikzlibrary{arrows.meta,backgrounds,calc,fit,positioning}
\usepackage{hyperref}

\title{Early Warning Signals for OpenVLA Failure under Visual Distribution Shift}

\author{
  Dipesh Tharu Mahato \\
  New York University \\
  \texttt{dm6259@nyu.edu} \\
  \And
  Rachel Ren \\
  New York University \\
  \texttt{rr4000@nyu.edu}
}

\begin{document}

\raggedbottom
\setlength{\textfloatsep}{10pt plus 2pt minus 2pt}
\setlength{\intextsep}{8pt plus 2pt minus 2pt}
\setlength{\floatsep}{10pt plus 2pt minus 2pt}
\makeatletter
\setlength{\@fptop}{0pt}
\setlength{\@fpsep}{10pt}
\makeatother

\maketitle

\begingroup
\renewcommand{\thefootnote}{}
\footnotetext{Code and experiment scripts are available at \url{https://github.com/dipeshbabu/vla-mech-monitor}.}
\addtocounter{footnote}{-1}
\endgroup

\begin{abstract}
Visual shifts can cause a vision-language-action policy to fail after initially plausible behavior. We ask whether OpenVLA's internal activations contain signals associated with the steps before failure. We freeze the policy, record one MLP activation per LIBERO-10 step, and fit two linear monitors. Occlusion reduces task success from $57\%$ to $17\%$. Within failed matched-reset trajectories, a layer-16 logistic probe attains AUROC $0.972$ and AUPRC $0.352$, whereas action disagreement attains AUROC $0.496$. Without refitting, the occlusion-trained probe reaches AUROC $0.689$ on failed camera-jitter episodes. In a calibration check, however, the same layer-16 monitor averages 3.32 warning onsets per clean episode. This contrast shows that strong retrospective discrimination does not imply operationally quiet warning behavior. Because fitting and evaluation share tasks, resets, and seed, these results establish retrospective separability rather than prediction on independent episodes.
\end{abstract}

\section{Introduction}
\label{sec:introduction}

Vision-language-action (VLA) models map camera observations and language instructions directly to robot actions. RT-1 and RT-2 established this interface at scale, while OpenVLA released a 7B-parameter policy and its training code \cite{brohan2022rt,zitkovich2023rt,kim2024openvla}. Because these policies place more of perception and control inside one learned model, an operator may have little external evidence that the policy has left its competence range. A rollout can begin with reasonable motion, drift after a visual change, and become unrecoverable before its actions look obviously wrong.

Runtime monitoring needs a useful signal before recovery becomes impossible. We ask whether internal policy states can provide that signal. Existing monitors use action uncertainty, predictive models, external vision-language models, or features from the policy itself \cite{liu2023runtime,xu2025can,duan2025aha,gu2025safe}. We study the last source with a linear readout attached to a frozen policy.

We use \emph{early warning} in a narrow empirical sense: a score associated with steps in a fixed window before a logged failure. Our question is whether a linear readout can recover that temporal signal. The readout may instead recognize the visual shift, identify a familiar reset, or exploit adjacent frames, so the evaluation design determines what its score means.

We test this distinction with the OpenVLA checkpoint fine-tuned on LIBERO-10 \cite{kim2024openvla,liu2023libero}. We change only the rendered image and record the input to an MLP down projection at each environment step. Occlusion is the primary stress test; camera jitter and channel-wise color shift test transfer. We compare a difference of episode means, a step-level logistic probe, and action disagreement under small input perturbations.

Fitting and evaluation reuse task IDs, reset identities, and seed; we call the resulting episodes \emph{matched-reset trajectories}. Occlusion lowers task success from $57\%$ to $17\%$. When ranking pre-failure-window steps within failed matched-reset trajectories, the layer-16 logistic probe reaches AUROC $0.972$ and AUPRC $0.352$, while action disagreement remains at chance. Without refitting, the probe retains weaker but above-chance ranking under camera jitter. Thresholding also exposes an operational gap: the strongest recorded probe still triggers 3.32 times per clean episode.

This paper makes three contributions:
\begin{itemize}
  \item It defines a controlled protocol for recording OpenVLA activations and assigning pre-failure-window labels under visual shift.
  \item It shows that selected feedforward states linearly separate pre-failure-window labels on matched-reset trajectories and outperform an action-disagreement baseline.
  \item It characterizes two limits of the signal: transfer weakens under camera jitter, and high ranking performance still produces frequent clean warnings.
\end{itemize}

The historical probe uses $K_{\mathrm{fit}}=20$, while evaluation uses $K_{\mathrm{eval}}=15$. We therefore interpret the reported metrics as retrospective separability rather than independent prediction. Figure~\ref{fig:architecture} states this evaluation boundary, and Table~\ref{tab:protocol} gives the precise protocol.

\section{Related Work}
\label{sec:related}

\newcommand{\relatedheading}[1]{\par\smallskip\noindent\textbf{#1}\ }

\relatedheading{Generalist robot policies.}
Generalist robot policies increasingly place perception, language grounding, and control inside learned models. CLIPort and BC-Z study language-conditioned manipulation and zero-shot imitation, while SayCan and PaLM-E connect language reasoning to learned affordances and sensor inputs \cite{shridhar2022cliport,jang2022bcz,ichter2023saycan,driess2023palme}. Gato, VIMA, RoboCat, and RoboFlamingo extend learned control across tasks, embodiments, or pretrained vision-language models \cite{reed2022gato,jiang2023vima,bousmalis2023robocat,li2023roboflamingo}. Open X-Embodiment, DROID, and Octo broaden the data and initialization available to these policies \cite{oneill2023openx,khazatsky2024droid,octo2024}. Action generation now spans autoregressive VLA models, diffusion policies, and flow-based policies \cite{brohan2022rt,zitkovich2023rt,kim2024openvla,chi2023diffusion,ze2024dp3,black2024pi0}. These architectures place more of the control pipeline inside a learned policy, which motivates monitoring its internal state.

LIBERO evaluates transfer across manipulation tasks and provides the simulator used here \cite{liu2023libero}. We use only five LIBERO-10 tasks, so this paper does not assess the broad task generalization targeted by the benchmark. The selected checkpoint and task subset instead provide a controlled setting in which the image can change while the simulator state and success rule remain fixed.

\relatedheading{Failure detection and uncertainty.}
Classical out-of-distribution detectors score novelty through softmax confidence or temperature-scaled input perturbations \cite{hendrycks2017baseline,liang2018odin}. In robotics, novelty alone is not the operational target: an unusual observation may be harmless, while a familiar observation may precede a task failure. Task-driven detectors therefore connect distribution shift or model error to downstream performance \cite{farid2022taskdriven,farid2023failure}. A system-level treatment also accounts for temporal dependence, closed-loop feedback, and the other components around a learned policy \cite{sinha2022system}.

Robot monitors now use several signal sources. Deep ensembles estimate predictive uncertainty but require multiple trained predictors \cite{lakshminarayanan2017ensembles}; predictive uncertainty can also degrade under dataset shift \cite{ovadia2019trust}. Model-based runtime monitoring rolls latent dynamics forward to anticipate risky states \cite{liu2023runtime}. KnowNo calibrates when a language-model planner should ask for help, and uncertainty-aware safety filters use conformal thresholds and latent dynamics to avoid unfamiliar hazards \cite{ren2023knowno,seo2025uncertainty}. Conformal prediction provides finite-sample coverage under exchangeability, but that guarantee does not automatically survive temporal dependence or a changed deployment distribution \cite{angelopoulos2021conformal}. We therefore describe our percentile threshold as a calibration rule, not a safety guarantee.

The closest manipulation studies differ in supervision and signal. FAIL-Detect constructs scalar uncertainty signals and fits a sequential detector without failure training data \cite{xu2025can}. SuccessVQA treats success detection as visual question answering with a separate pretrained vision-language model \cite{du2023success}; AHA uses another external model to recognize and explain failures \cite{duan2025aha}. SAFE trains a scalar detector from VLA features and tests unseen-task transfer \cite{gu2025safe}. Test-time policy adaptation extends failure detection by incorporating diagnosis and corrective feedback \cite{peng2023diagnosis}.

SAFE is the closest feature-based comparison, but our scientific target differs. We define a temporally localized pre-failure-window label under controlled visual shifts, then test cross-shift transfer and measure warning frequency after thresholding. We use the linear monitor to study pre-failure representations rather than to build a complete failure-detection system. The main analysis remains passive; Appendix~\ref{app:warning} reports the exploratory response policies separately.

\relatedheading{Probing and mechanistic claims.}
Linear probes measure whether a target is accessible in a representation \cite{alain2016understanding}. Layer-wise probing has also traced linguistic structure across pretrained transformer depth \cite{tenney2019bert}. Reviews of representation analysis stress that high probe accuracy alone says little about how a model uses the decoded information \cite{belinkov2019analysis}. Probe capacity, correlated examples, and control tasks can change the interpretation \cite{hewitt2019control}; information-theoretic accounts make the target and available data equally important \cite{pimentel2020information}. Iterative nullspace projection removes linearly decodable attributes \cite{ravfogel2020null}; amnesic probing then tests how such removal changes model behavior \cite{elazar2021amnesic}.

Recent VLA work has begun to make such interventions. H{\"a}on et al. identify feedforward directions and inject them to steer actions, while Khan et al. steer a VLA through sparse latent directions \cite{haon2025mechanistic,khan2025sparse}. We record a related feedforward site, but never alter it. Accordingly, a high score in our experiment means only that the label is linearly decodable under the stated split. It does not locate a mechanism of failure.

\relatedheading{Visual distribution shift.}
Common-corruption benchmarks show that small changes in image formation can expose brittle visual features \cite{hendrycks2019corruptions}. Recreated ImageNet test sets and WILDS document accuracy loss under new data collection and naturally occurring shifts \cite{recht2019imagenet,koh2021wilds}. Shortcut learning gives a useful warning for this study: a classifier may perform well by detecting a condition that correlates with the label in one dataset \cite{geirhos2020shortcut}. Domain randomization addresses the complementary training problem by varying simulated appearance to improve transfer \cite{tobin2017domain}. We do not retrain or robustify OpenVLA. Instead, paired visual shifts serve as controlled stress tests, and held-out shifts are necessary to determine whether the monitor reads impending task failure or merely recognizes the perturbation used during fitting.

\section{Method}
\label{sec:method}

\begin{figure}[!t]
  \centering
  \resizebox{\textwidth}{!}{\begin{tikzpicture}[
  font=\sffamily,
  >={Stealth[length=2.0mm]},
  online/.style={->, line width=0.95pt, draw=blue!62!black},
  recorded/.style={->, dashed, line width=0.95pt, draw=green!45!black},
  recordedline/.style={dashed, line width=0.95pt, draw=green!45!black},
  captureflow/.style={->, dashed, line width=0.85pt, draw=red!72!black},
  feedback/.style={->, line width=0.85pt, draw=black!58},
  panel/.style={draw=black!34, rounded corners=5pt, fill=black!1,
                line width=0.75pt},
  base/.style={draw=black!40, rounded corners=2pt, fill=white,
               align=center, font=\footnotesize, inner xsep=2.5pt,
               inner ysep=2.5pt, minimum height=0.76cm},
  model/.style={base, draw=blue!65!black, fill=blue!7},
  shiftbox/.style={base, draw=orange!80!black, fill=orange!10},
  hook/.style={base, draw=red!72!black, fill=red!8, line width=1.0pt,
               font=\scriptsize, minimum height=1.35cm},
  data/.style={base, draw=green!48!black, fill=green!8,
               font=\scriptsize, minimum height=1.18cm},
  monitor/.style={base, draw=green!48!black, fill=green!8,
                  font=\scriptsize, minimum height=0.88cm},
  eval/.style={base, draw=violet!66!black, fill=violet!8,
               font=\scriptsize, minimum height=0.94cm},
  layer/.style={draw=blue!60!black, rounded corners=1pt, fill=blue!5,
                minimum width=0.48cm, minimum height=0.76cm,
                font=\scriptsize, align=center, inner sep=1pt},
  fusion/.style={draw=blue!65!black, circle, fill=blue!7,
                 minimum size=0.40cm, font=\scriptsize\bfseries, inner sep=0pt},
  picked/.style={layer, draw=red!76!black, fill=red!13, line width=1.0pt},
  title/.style={font=\footnotesize\bfseries, text=black!76},
  note/.style={font=\scriptsize, text=black!62, align=center}
]

\node[panel, minimum width=4.20cm, minimum height=3.60cm]
  (pobs) at (2.10,6.95) {};
\node[panel, minimum width=8.87cm, minimum height=3.60cm]
  (pmodel) at (8.935,6.95) {};
\node[panel, minimum width=3.58cm, minimum height=3.60cm]
  (pcontrol) at (15.46,6.95) {};

\node[panel, minimum width=5.00cm, minimum height=2.80cm]
  (pcapture) at (2.50,2.70) {};
\node[panel, minimum width=5.60cm, minimum height=2.80cm]
  (plabel) at (8.10,2.70) {};
\node[panel, minimum width=6.05cm, minimum height=2.80cm]
  (pfit) at (14.225,2.70) {};

\node[title, anchor=west] at (0.00,9.08) {ONLINE CLOSED LOOP: FROZEN OPENVLA};
\node[title] at (2.10,8.10) {1. Form policy inputs};
\node[title] at (8.935,8.10) {2. Run the multimodal policy};
\node[title] at (15.46,8.10) {3. Execute the action};

\node[base, text width=0.80cm] (state) at (0.66,7.08)
  {state\\$x_{e,t}$};
\node[base, text width=0.80cm] (render) at (1.94,7.08)
  {render\\$o_{e,t}$};
\node[shiftbox, text width=1.12cm] (perturb) at (3.38,7.08)
  {$q_\rho(o_{e,t})$\\shift pixels};
\draw[online] (state) -- (render);
\draw[online] (render) -- (perturb);
\node[model, font=\scriptsize, text width=1.36cm] (instruction) at (2.10,5.83)
  {instruction $c_e$\\tokenizer};

\node[model, font=\scriptsize, text width=1.40cm] (vision) at (5.53,7.08)
  {DINOv2 +\\SigLIP\\visual towers};
\node[model, font=\scriptsize, text width=0.94cm] (projector) at (7.08,7.08)
  {visual\\projector};
\node[fusion] (fusion) at (8.12,7.08) {$+$};
\node[layer] (l0) at (8.92,7.08) {$L_0$};
\node[layer] (ld) at (9.76,7.08) {$\cdots$};
\node[picked] (lL) at (10.60,7.08) {$L$};
\node[layer, minimum width=0.54cm] (l31) at (11.47,7.08) {$L_{31}$};
\node[model, font=\scriptsize, text width=0.78cm] (tokens) at (12.58,7.08)
  {action\\tokens};
\draw[online] (perturb) -- (vision);
\draw[online] (vision) -- (projector);
\draw[online] (projector) -- (fusion);
\draw[online] (fusion) -- (l0);
\draw[online] (l0) -- (ld);
\draw[online] (ld) -- (lL);
\draw[online] (lL) -- (l31);
\draw[online] (l31) -- (tokens);
\draw[online] (instruction.east) -- (4.35,5.83) -- (8.12,5.83)
  -- (fusion.south);
\node[note] at (10.60,7.66)
  {record one of 32 Llama~2 blocks per rollout};

\node[model, text width=1.38cm] (detok) at (14.62,7.08)
  {detokenize\\$a_{e,t}\in\mathbb{R}^{7}$};
\node[base, text width=1.22cm] (sim) at (16.34,7.08)
  {LIBERO\\transition};
\draw[online] (tokens) -- (detok);
\draw[online] (detok) -- (sim);
\node[note] at (15.46,5.83)
  {execute $a_{e,t}$ and obtain $x_{e,t+1}$};
\draw[feedback] (sim.east) -- (17.55,7.08) -- (17.55,9.68) -- (-0.28,9.68)
  -- (-0.28,7.08) -- (state.west);
\node[note, anchor=east] at (16.10,9.84) {next environment step};

\node[title, anchor=west] at (0.00,4.62)
  {RECORDED DATA: OFFLINE FITTING AND EVALUATION};
\node[title] at (2.50,3.76) {4. Capture one feedforward state};
\node[title] at (8.10,3.76) {5. Construct labels and partitions};
\node[title] at (14.225,3.76) {6. Fit and evaluate each monitor};

\node[hook, text width=1.38cm] (site) at (1.08,2.63)
  {pre-hook at\\\texttt{down\_proj}\\passes $1,\ldots,P$};
\node[hook, text width=2.28cm] (pool) at (3.47,2.63)
  {retain pass $P$\\$A^{(L,P)}_{e,t}\in\mathbb{R}^{q\times d}$\\
   $h^{(L)}_{e,t}=q^{-1}\!\sum_j A^{(L,P)}_{e,t,j}$};
\draw[captureflow] (lL.south) -- (10.60,4.34) -- (-0.28,4.34)
  -- (-0.28,2.63) -- (site.west);
\draw[recorded] (site) -- (pool);
\node[note] at (2.50,1.49)
  {usually $q=1$: the final action-token state};

\node[data, text width=1.02cm] (events) at (6.20,2.63)
  {episode\\outcome +\\first event $\tau_e$};
\node[data, text width=1.48cm] (labels) at (8.01,2.63)
  {target $y_{e,t}=1$\\iff $0\leq\tau_e-t\leq K$\\else $0$};
\node[data, text width=1.20cm] (partition) at (9.91,2.63)
  {group by\\episode, seed,\\task, shift};
\draw[recorded] (pool) -- (events);
\draw[recorded] (events) -- (labels);
\draw[recorded] (labels) -- (partition);
\node[note] at (8.10,1.49)
  {fit preprocessing statistics on training groups only};

\node[monitor, text width=1.18cm] (direction) at (12.85,3.08)
  {mean-direction\\monitor};
\node[monitor, text width=1.18cm] (logistic) at (12.85,2.08)
  {logistic\\probe};
\node[eval, text width=2.15cm] (metrics) at (15.64,2.58)
  {AUROC / AUPRC\\lead time / false alarms};
\coordinate (split) at (11.75,2.63);
\coordinate (merge) at (14.00,2.58);
\draw[recordedline] (partition.east) -- (split);
\draw[recordedline] (11.75,2.08) -- (11.75,3.08);
\draw[recorded] (11.75,3.08) -- (direction.west);
\draw[recorded] (11.75,2.08) -- (logistic.west);
\draw[recordedline] (direction.east) -- (14.00,3.08);
\draw[recordedline] (logistic.east) -- (14.00,2.08);
\draw[recordedline] (14.00,2.08) -- (14.00,3.08);
\draw[recorded] (merge) -- (metrics.west);
\node[note] at (14.225,1.49)
  {score evaluation groups separately\\at each selected layer $L$};

\node[draw=blue!58!black, fill=blue!4, rounded corners=2pt,
      text width=8.05cm, minimum height=0.58cm, inner xsep=2pt,
      align=center, font=\scriptsize] at (4.225,0.70)
  {\textbf{EVALUATION BOUNDARY: CURRENT STUDY}\\
   seed 7; matched-reset trajectories; $K_{\rm fit}=20$, $K_{\rm eval}=15$};
\node[draw=green!45!black, fill=green!5, rounded corners=2pt,
      text width=8.05cm, minimum height=0.58cm, inner xsep=2pt,
      align=center, font=\scriptsize] at (13.00,0.70)
  {\textbf{EVALUATION BOUNDARY: DEPLOYMENT VALIDATION}\\
   disjoint resets/seeds; fixed $K$; calibrated threshold; validated supervisor};

\node[note, anchor=west] at (0.00,0.10)
  {solid blue: online execution \quad dashed green: recorded/offline flow
   \quad dashed red: activation capture};

\end{tikzpicture}}
  \caption{OpenVLA rollout and monitoring pipeline. Blue arrows trace one closed-loop policy step, green dashed arrows trace the offline fitting path, and the red hook records the input to one MLP down projection. The bottom row separates the current analysis of matched-reset trajectories from deployment validation.}
  \label{fig:architecture}
\end{figure}
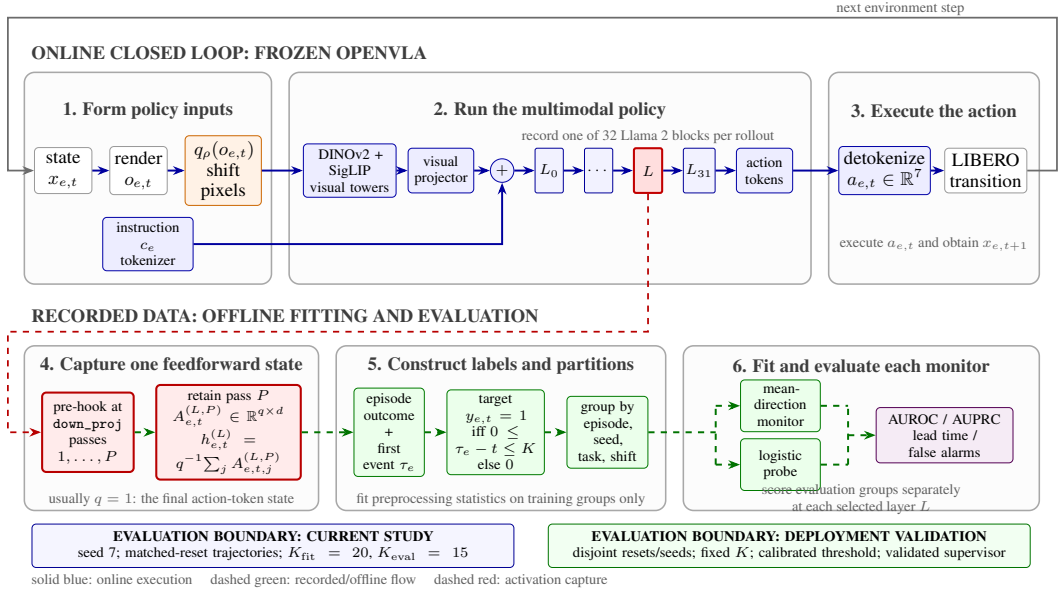

Figure~\ref{fig:architecture} summarizes the pipeline. OpenVLA acts on a perturbed image while a hook records one feedforward state per step. We fit the monitors after rollout collection and never update the policy. The logger records one layer at a time, so each layer requires a separate rollout.

\subsection{From visual shift to monitor score}
\label{sec:target}

Let $e$ index an episode and $t$ an environment step. The simulator has state $x_{e,t}$, renders an RGB observation $o_{e,t}$, and supplies a fixed language instruction $c_e$. A transformation $q_\rho$ produces the policy image $\tilde{o}_{e,t}=q_\rho(o_{e,t})$, where $\rho$ selects the perturbation type and strength. The frozen policy then executes
\begin{equation}
  a_{e,t}=\pi_\theta(\tilde{o}_{e,t},c_e),
  \qquad x_{e,t+1}\sim P(\cdot\mid x_{e,t},a_{e,t}).
\end{equation}
The transformation leaves $x_{e,t}$, the instruction, the simulator dynamics $P$, and the success rule unchanged. Later states can still diverge because the policy acts on a changed image.

The monitor maps layer-$L$ activation $h_{e,t}^{(L)}$ to a scalar score $r_{e,t}^{(L)}$. We test whether this score ranks steps near a logged failure above negative steps. The recorded experiment is therefore a retrospective ranking test; Section~\ref{sec:discussion} states the additional requirements for early prediction.

\subsection{Capturing one activation per step}
\label{sec:capture}

OpenVLA combines DINOv2 and SigLIP visual features, projects them into the language model, and autoregressively emits action tokens \cite{oquab2023dinov2,zhai2023siglip,kim2024openvla}. A forward pre-hook on \texttt{mlp.down\_proj} captures the expanded MLP state before the module maps it back to the residual width.

Autoregressive generation calls this module several times per action. Let $A_{e,t}^{(L,p)}\in\mathbb{R}^{q_p\times d_L}$ denote its output on decoding pass $p$. The callback overwrites its buffer on every pass, retains the final output $A_{e,t}^{(L,P)}$, and averages any remaining sequence positions:
\begin{equation}
  h_{e,t}^{(L)}=\frac{1}{q_P}\sum_{j=1}^{q_P} A_{e,t,j}^{(L,P)}.
  \label{eq:activation}
\end{equation}
Thus, each environment step contributes one activation vector from the final autoregressive action-token pass. Usually $q_P=1$, so averaging leaves a single token state rather than combining image and instruction tokens. Appendix~\ref{app:implementation} gives the implementation details.

\subsection{Defining the failure window}
\label{sec:labels}

Each episode has a simulator success flag and zero or more detector events. For a failed episode, the logger sets $\tau_e$ to the first wrong-object, drop, or stalled-progress event. If none fires, it uses the terminal timeout. The logger discards provisional events from episodes that later succeed. Table~\ref{tab:failure_taxonomy} defines these unaudited events, and Appendix~\ref{app:logger} describes their implementation.

For horizon $K$, we define the pre-failure-window label for a failed episode as
\begin{equation}
  y_{e,t}^{(K)}=\mathbf{1}\{0\leq \tau_e-t\leq K\}.
  \label{eq:label}
\end{equation}
All steps from successful episodes are negative. Within failed episodes, the logistic fitter uses steps more than $3K$ before $\tau_e$ as negatives and omits the intervening band. This gap reduces label ambiguity near the window but cannot correct an inaccurate $\tau_e$ or remove temporal dependence.

\subsection{Linear monitors and action baseline}
\label{sec:monitors}

\paragraph{Mean-difference direction.}
We first average activations over time within each episode, so long episodes do not contribute more vectors than short episodes. Let $\mu_{\mathrm{fail}}^{(L)}$ and $\mu_{\mathrm{succ}}^{(L)}$ denote the means of those episode vectors. We form the unit direction
\begin{equation}
  \hat v^{(L)}=
  \frac{\mu_{\mathrm{fail}}^{(L)}-\mu_{\mathrm{succ}}^{(L)}}
       {\lVert\mu_{\mathrm{fail}}^{(L)}-\mu_{\mathrm{succ}}^{(L)}\rVert_2}
\end{equation}
and score a step with $r_{\mathrm{dir}}(e,t,L)=\langle h_{e,t}^{(L)},\hat v^{(L)}\rangle$. This readout tests a simple geometric hypothesis: failed and successful episodes differ along one unregularized mean direction.

\paragraph{Logistic probe.}
The second monitor fits logistic regression to the pre-failure-window labels in Equation~\ref{eq:label}. Statistics from the fitting subset standardize each activation dimension. The score is
\begin{equation}
  r_{\mathrm{log}}(e,t,L)=
  \sigma\!\left(w_L^\top\frac{h_{e,t}^{(L)}-m_L}{s_L}+b_L\right),
  \label{eq:logistic}
\end{equation}
where $m_L$ and $s_L$ come from the fitting partition. The fitter caps the negative-to-positive ratio before optimization. Unlike the mean direction, this probe learns a boundary for the step-level target. Appendix~\ref{app:probe} gives the fitting details and documents the historical horizon mismatch.

\paragraph{Action disagreement.}
The baseline does not read model activations. It applies small random image jitters to one observation, samples the resulting OpenVLA actions, and averages their variance across action dimensions. This score asks whether local action instability supplies the same warning without instrumenting the network. It is a perturbation-sensitivity baseline rather than a calibrated estimate of epistemic uncertainty.

\subsection{Metrics and warning threshold}
\label{sec:evaluation}

We report AUROC and AUPRC over labeled steps. AUROC measures ranking across positive and negative steps; AUPRC is more sensitive to the rare positive class, and its random-ranking reference equals the positive-step prevalence. The current tables contain point estimates. A held-out analysis must place each episode in one partition and resample whole episodes for confidence intervals because adjacent steps share the same rollout.

For warning analysis, we set the threshold to the 95th percentile of scores on a clean calibration run. We report warning onsets per episode, active-warning time, and lead time to $\tau_e$. Table~\ref{tab:clean_warning} reuses the calibration run, so it is a calibration check rather than a held-out false-alarm estimate. Appendix~\ref{app:warning} gives the timing logic and exploratory response policies.

\section{Experiments}
\label{sec:experiments}

\subsection{Setup}

We evaluate \texttt{openvla/openvla-7b-finetuned-libero-10} on LIBERO-10 task IDs 0--4. Each condition contains 20 trials per task under seed 7, for 100 episodes. For a given task and episode index, fitting and evaluation call \texttt{set\_init\_state} with the same stored LIBERO state. Paired episodes therefore start from identical simulator states, although their trajectories can diverge after the visual shift changes the policy's actions. At each step, we store the task, episode, action, success flag, detector events, selected activation, and warning state. We compute action disagreement only for its baseline run.

\subsection{Visual shifts and transfer}

The clean condition leaves the image unchanged. Occlusion adds a black rectangle, camera jitter translates the image and zero-fills exposed pixels, and color shift rescales the RGB channels. Every transformation uses strength 0.35. We fit both activation monitors on occlusion, then apply them to camera jitter and color shift without refitting. Task, reset, and seed remain matched in this transfer test.

\subsection{Evaluation scope}

Table~\ref{tab:protocol} summarizes the recorded protocol and the split needed to estimate generalization.

\begin{table}[t]
  \centering
  \footnotesize
  \setlength{\tabcolsep}{4pt}
  \caption{Recorded evidence versus a held-out protocol. The current runs do not test unseen episodes.}
  \label{tab:protocol}
  \begin{tabular}{@{}>{\raggedright\arraybackslash}p{0.18\linewidth}>{\raggedright\arraybackslash}p{0.34\linewidth}>{\raggedright\arraybackslash}p{0.39\linewidth}@{}}
    \toprule
    Aspect & Recorded evidence & Held-out protocol \\
    \midrule
    Coverage & Tasks 0--4; 20 trials per task and condition & Predeclared tasks across multiple disjoint seeds \\
    Split & Separate runs reuse task, reset, and seed & Disjoint reset identities; separate seed, task, and shift tests \\
    Layers & Blocks 8, 10, and 16 & Predeclared sweep over all 32 blocks \\
    Target & Unaudited events; $K_{\rm fit}=20$, $K_{\rm eval}=15$ & Audited events; one fixed $K$ plus horizon sensitivity \\
    Shifts & Occlusion fit; jitter and color transfer at strength 0.35 & Strength chosen on calibration data; mixed outcomes on test data \\
    Uncertainty & Point estimates over correlated steps & Episode bootstrap with task- and seed-level summaries \\
    \bottomrule
  \end{tabular}
\end{table}

We first measure task success because a detection metric is undefined when a condition produces only successes or only failures. The primary contrast ranks positive and negative steps within failed occluded rollouts. We test both activation monitors at layers 8, 10, and 16 and compare them with action disagreement. The historical exploratory runs retained these layers without a documented prespecified selection rule, so we report all three and do not treat layer 16 as independently selected. This sweep covers only three of OpenVLA's 32 language-model blocks.

For transfer, we apply the layer-16 occlusion monitors to camera jitter and color shift. Table~\ref{tab:transfer} includes only failed episodes. We report the all-episode analysis separately because it adds successful episodes to the negative class and changes the comparison.

We evaluate against the $K=15$ pre-failure-window label with AUROC and AUPRC. Because the historical fit used $K=20$, we interpret its metrics as separability under nearby temporal targets rather than as a calibrated estimate for one fixed horizon. The logger stores per-step activations and failure times, so retained traces permit offline refitting at $K=15$ without new OpenVLA rollouts.

We also report warning onsets per episode and the fraction of steps under an active warning. Ranking metrics alone do not show how often a threshold interrupts clean behavior.

\section{Results}
\label{sec:results}

\subsection{Occlusion reduces success by 40 points}

OpenVLA succeeds in 57 of 100 clean episodes and 17 of 100 occluded episodes (Table~\ref{tab:success}). Thus, the selected occlusion disrupts task execution without making failure equivalent to the occlusion condition. Because all runs use seed 7, these proportions describe this collection rather than average LIBERO-10 performance.

\begin{table}[H]
  \centering
  \small
  \caption{Task success under seed 7. Each condition contains 100 episodes across five LIBERO-10 tasks.}
  \label{tab:success}
  \begin{tabular}{@{}lcc@{}}
    \toprule
    Condition & Successful episodes & Success rate \\
    \midrule
    Clean & 57/100 & 57\% \\
    Occlusion & 17/100 & 17\% \\
    \bottomrule
  \end{tabular}
\end{table}

\subsection{Pre-failure-window labels are linearly accessible in selected feedforward states}

\begin{figure}[H]
  \centering
  \includegraphics[width=\linewidth]{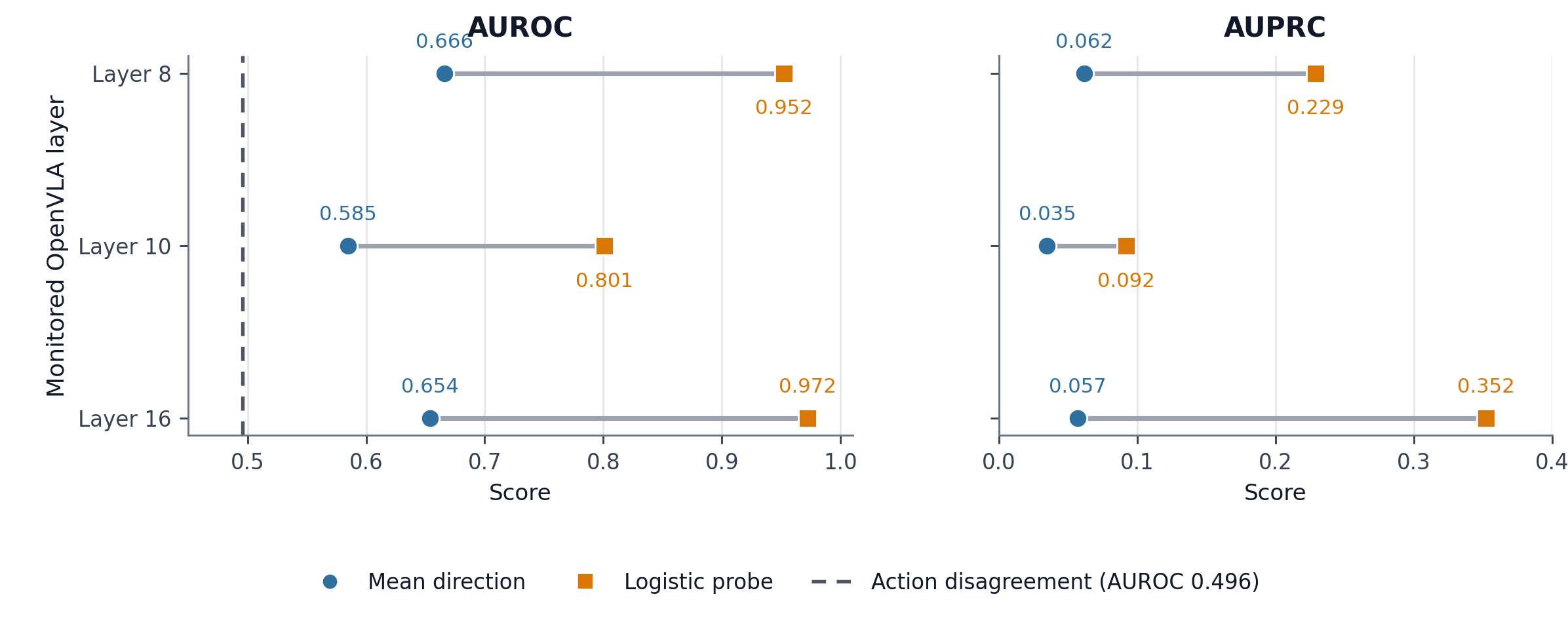}
  \caption{Layer-wise discrimination under occlusion. Each panel reports one metric, so AUROC and AUPRC retain separate scales. Circles show the mean-direction monitor; squares show logistic regression. Horizontal segments show the difference between monitors at each layer. The dashed AUROC line marks the action-disagreement baseline ($0.496$). The matched-reset trajectories share task, reset, and seed.}
  \label{fig:occluded_layer_performance}
\end{figure}

Across the three sampled layers, logistic readouts reveal substantial linear accessibility of the pre-failure-window label. The strongest recorded discrimination occurs at layer 16, with AUROC $0.972$ and AUPRC $0.352$ against the 15-step label (Figure~\ref{fig:occluded_layer_performance} and Table~\ref{tab:layer_performance}). At the same layer, the mean direction reaches AUROC $0.654$ and AUPRC $0.057$. Action disagreement reaches AUROC $0.496$.

\begin{table}[H]
  \centering
  \small
  \caption{Ranking over steps from failed occluded episodes. Fit and evaluation share task, reset, and seed; the logistic probe also uses $K_{\rm fit}=20$ and $K_{\rm eval}=15$.}
  \label{tab:layer_performance}
  \begin{tabular}{@{}llccc@{}}
    \toprule
    Monitor & Metric & Layer 8 & Layer 10 & Layer 16 \\
    \midrule
    Mean direction & AUROC & 0.666 & 0.585 & 0.654 \\
                   & AUPRC & 0.062 & 0.035 & 0.057 \\
    \addlinespace
    Logistic probe & AUROC & 0.952 & 0.801 & 0.972 \\
                   & AUPRC & 0.229 & 0.092 & 0.352 \\
    \bottomrule
  \end{tabular}
\end{table}

Performance does not vary monotonically across the three sampled layers. Logistic AUROC falls from $0.952$ at layer 8 to $0.801$ at layer 10, then rises to $0.972$ at layer 16; the mean direction follows a different order. Because the sweep omits 29 of 32 blocks, it cannot locate where the signal emerges or select an optimal layer.

\subsection{Camera-jitter transfer remains above chance}

On failed camera-jitter episodes, the layer-16 direction reaches AUROC $0.635$ and the logistic probe reaches $0.689$ without refitting (Table~\ref{tab:transfer}). Both exceed chance but trail their occlusion scores. This transfer weakens the narrowest shortcut account in which the probe responds only to the occlusion artifact; it does not rule out a shared image-degradation or task-progress feature. Including successful episodes changes the negative class and yields AUROC $0.731$ and $0.725$, respectively.

Color shift causes no failures in 100 episodes. Equation~\ref{eq:label} therefore produces no positives, and AUROC/AUPRC are undefined. This condition cannot test detection at the selected strength.

\begin{table}[H]
  \centering
  \footnotesize
  \setlength{\tabcolsep}{2.6pt}
  \caption{Transfer on steps from failed episodes. Occlusion data fitted both monitors; jitter and color use no refitting. A dash marks an undefined metric because color shift produced no failed episodes.}
  \label{tab:transfer}
  \begin{tabular}{@{}lcccc@{}}
    \toprule
    & \multicolumn{2}{c}{Direction} & \multicolumn{2}{c}{Logistic} \\
    \cmidrule(lr){2-3}\cmidrule(l){4-5}
    Shift & AUROC & AUPRC & AUROC & AUPRC \\
    \midrule
    Occlusion fit & 0.654 & 0.057 & 0.972 & 0.352 \\
    Camera jitter & 0.635 & 0.051 & 0.689 & 0.062 \\
    Color shift & -- & -- & -- & -- \\
    \bottomrule
  \end{tabular}
\end{table}

\subsection{High discrimination still yields frequent clean warnings}

Under the no-op response, which substitutes LIBERO's dummy action while a warning is active, clean runs produce 3.32--6.05 warning onsets per episode, and warnings are active for $2.63\%$--$4.92\%$ of clean steps (Table~\ref{tab:clean_warning}). The layer-16 logistic probe has the lowest active fraction but still produces 3.32 warning onsets per episode. Strong retrospective discrimination therefore does not imply operationally quiet warning behavior.

\begin{table}[H]
  \centering
  \footnotesize
  \setlength{\tabcolsep}{2.7pt}
  \caption{Warnings on clean no-op rollouts. ``Onsets/ep'' counts warning onsets per episode; ``Active'' gives the percentage of warned steps. The threshold uses these same rollouts, so the values are calibration checks.}
  \label{tab:clean_warning}
  \begin{tabular}{@{}lcccc@{}}
    \toprule
    & \multicolumn{2}{c}{Direction} & \multicolumn{2}{c}{Logistic} \\
    \cmidrule(lr){2-3}\cmidrule(l){4-5}
    Layer & Onsets/ep & Active & Onsets/ep & Active \\
    \midrule
    L8  & 4.05 & 3.25\% & 3.97 & 3.06\% \\
    L10 & 6.05 & 4.92\% & 4.32 & 3.24\% \\
    L16 & 5.19 & 4.08\% & 3.32 & \textbf{2.63\%} \\
    \bottomrule
  \end{tabular}
\end{table}

These counts reuse the threshold's calibration distribution. Appendix~\ref{app:warning} reports the response-policy grids in Tables~\ref{tab:direction_warnings} and~\ref{tab:logistic_warnings}. We keep those results out of the main comparison because aborting changes the trajectory used by passive AUROC.

\section{Safety Interpretation and Limitations}
\label{sec:discussion}

\subsection{What the internal signal contributes}

The results support a narrow claim: selected OpenVLA feedforward states contain linearly accessible information associated with the pre-failure window under the recorded protocol. Within failed matched-reset trajectories, the classifier outperforms action disagreement, but this comparison does not show that activations add information unavailable from pixels or other action-level signals. Observation-only and external success baselines would test that stronger claim.

The camera-jitter result narrows one alternative explanation. Transfer without refitting argues against a detector specific to the occlusion artifact, though the probe may still read a generic image-quality or task-progress cue.

The gap between monitors also has a limited interpretation. A fitted step-level hyperplane separates the labels more strongly than the raw difference between successful and failed episode means. This boundary may reflect a failure-related feature, a nuisance variable, or the monitors' different sampling units. It does not show that OpenVLA uses the feature to choose actions.

\subsection{Threats to validity}

\paragraph{Matched-reset trajectories.}
Deterministic resets pair fitting and evaluation by task, object layout, and initial state. Step-wise metrics amplify this dependence, and grouping by episode cannot fix a second run that recreates the same episode. A valid outer split must hold out reset identities or seeds.

\paragraph{Failure labels.}
Stalled progress relies on a heuristic, and a timeout may occur well after recovery became impossible. The logger also removes events from episodes that eventually succeed, so future outcome affects the label. A manual audit must measure event precision, coverage, and timing before a horizon study can establish useful lead time. Table~\ref{tab:failure_taxonomy} summarizes the current rules.

\begin{table}[H]
  \centering
  \scriptsize
  \setlength{\tabcolsep}{3pt}
  \caption{Operational failure labels. The event times are unaudited heuristics, so a manual label audit remains necessary. Distances use simulator units.}
  \label{tab:failure_taxonomy}
  \begin{tabular}{@{}>{\raggedright\arraybackslash}p{0.18\linewidth}>{\raggedright\arraybackslash}p{0.56\linewidth}>{\raggedright\arraybackslash}p{0.20\linewidth}@{}}
    \toprule
    Category & Operational trigger & Recorded event \\
    \midrule
    Wrong object & Nearby object after gripper closure differs from the instruction-matched target. & \texttt{wrong\_object} \\
    Drop & Held object falls by more than $0.06$ and separates from the end effector by more than $0.12$. & \texttt{drop} \\
    Stalled progress & Target-distance improvement is less than $0.015$ over 25 steps. & \texttt{goal\_drift} \\
    Timeout & No heuristic fires before a failed rollout ends. & Terminal fallback \\
    \bottomrule
  \end{tabular}
\end{table}

\paragraph{Scope.}
The evidence covers five simulated tasks, one checkpoint, one seed, three visual transformations, and three internal layers. Even a successful held-out LIBERO result would not establish transfer to physical sensing, other embodiments, or VLA policies in general.

\subsection{Role in a runtime safety system}

An internal monitor could run beside a frozen policy and send its score to a separate supervisor. That supervisor might request help or transfer control to a validated stopping policy. We evaluate only score construction and retrospective ranking; we do not test a supervisor, a fallback policy, or a safety guarantee.

A deployment evaluation must use independent resets and seeds, calibrate thresholds on separate clean data, and measure false alarms, misses, and usable lead time. Table~\ref{tab:clean_warning} shows why ranking alone is insufficient. Repeating an action, sending a dummy action, or aborting does not diagnose the scene or recover the task. Appendix~\ref{app:warning} retains these exploratory responses only to document the completed runs.

Probe accuracy shows accessibility rather than mechanism \cite{hewitt2019control,elazar2021amnesic}. A causal test must intervene on a feature found on training tasks, change held-out actions and outcomes in the predicted direction, and pass random and norm-matched controls. VLA steering studies illustrate this design \cite{haon2025mechanistic,khan2025sparse}.

\paragraph{Future validation.}
A stronger validation protocol would first audit failure times, fix one horizon, and evaluate the selected layers on disjoint resets and seeds. Observation-only and external success baselines would test whether activations add information beyond observable inputs, while a full-depth sweep would test where that information appears. Causal intervention and supervisory control remain separate questions.

\section{Conclusion}
\label{sec:conclusion}

These experiments identify linearly accessible information associated with the pre-failure window in selected OpenVLA feedforward states under controlled visual shift. The signal is strong within failed matched-reset trajectories, transfers more weakly to camera jitter, and can still generate frequent warnings on clean rollouts. The experiments support internal state as a candidate monitoring signal, but they also show that representation-level separability does not yield a deployment-ready predictor. Evaluating such a predictor requires independent resets and seeds, a fixed temporal target, and separately calibrated thresholds. A safety response would also require a validated supervisor, while a mechanistic claim would require controlled interventions that change actions and outcomes.

\clearpage
\bibliographystyle{unsrtnat}
\bibliography{template}

\clearpage
\appendix
\section{Implementation and Supplementary Warning Results}
\label{app:implementation}

\subsection{Rollout logging and failure times}
\label{app:logger}

The experiments run OpenVLA on LIBERO-10 task IDs 0--4 with 20 trials per task, seed 7, and perturbation strength 0.35. Each condition therefore contains 100 episodes. The visual perturbations depend deterministically on the global seed, task ID, episode ID, and time step. Occlusion applies a black rectangle, color shift scales the RGB channels, and camera jitter translates the image while filling exposed pixels with zeros. These operations leave the instruction, simulator state, dynamics, and success rule unchanged at the moment of application.

The rollout logger records the first heuristic event associated with a wrong-object grasp, a drop, or stalled progress. If no event fires in a failed episode, the logger assigns the terminal timeout as the failure time. It removes detector events from episodes that later succeed. These event times provide monitoring labels, not manually verified semantic annotations.

\subsection{Probe fitting and evaluation}
\label{app:probe}

For the mean-difference monitor, the fitter first averages activations within each episode. It then computes separate means for successful and failed episodes and normalizes their difference. The monitor scores each new activation by its signed projection onto that unit vector.

For the logistic probe, steps no more than $K$ steps before the logged failure are positive. By default, failed-episode steps more than $3K$ steps before failure and all steps from successful episodes are negative; the fitter omits the intervening band. It caps the class ratio, standardizes activations on the fitting subset, and stores the normalization statistics with the probe. The recorded fit used the runner's default $K=20$, whereas the plotted evaluation label used $K=15$. Because the logger stores activations and failure times for every step, retained traces can repair this mismatch offline; new policy rollouts are unnecessary. Any replication should fix $K$ before fitting and evaluation.

The action-disagreement baseline does not use activations. It samples actions from several image-jittered copies of one observation and reports the mean variance across action dimensions. The evaluator applies the same step labels used for the activation monitors.

\subsection{Warning controller and intervention grids}
\label{app:warning}

The controller sets its threshold to the 95th percentile of scores from a clean calibration run. A warning begins after two consecutive threshold crossings, remains active for three steps, and then enters a five-step cooldown. The no-op policy substitutes LIBERO's dummy action, the abort policy ends the rollout, and the hold-last policy repeats the previous action. These policies test whether a score can drive a simple runtime response; they are not recovery controllers.

Tables~\ref{tab:direction_warnings} and~\ref{tab:logistic_warnings} give the full warning-policy results. Their baseline rows reproduce Table~\ref{tab:layer_performance}; Table~\ref{tab:clean_warning} reports clean calibration separately. Lead time and onset count do not apply to baselines. Because abort changes the trajectory, its zero-by-construction lead time and AUROC are not comparable to passive monitoring. A dash marks a missing value rather than zero.

\begin{table}[H]
\centering
\scriptsize
\setlength{\tabcolsep}{3.2pt}
\caption{Direction-monitor warnings under occlusion; abort metrics are descriptive only.}
  \label{tab:direction_warnings}
  \begin{tabular}{@{}llcccc@{}}
    \toprule
    Policy & Layer & AUROC & AUPRC & Lead time & Onsets/ep \\
    \midrule
    Baseline & L8  & 0.666 & 0.062 & -- & -- \\
             & L10 & 0.585 & 0.035 & -- & -- \\
             & L16 & 0.654 & 0.057 & -- & -- \\
    \midrule
    No-op & L8  & 0.604 & 0.037 & 405 & 6.99 \\
          & L10 & 0.572 & 0.034 & 430 & -- \\
          & L16 & 0.612 & 0.043 & 375 & 6.48 \\
    \midrule
    Abort & L8  & 0.772 & 0.247 & 0 & 0.48 \\
          & L10 & 0.674 & 0.102 & 0 & 0.29 \\
          & L16 & 0.731 & 0.153 & 0 & 0.55 \\
    \midrule
    Hold last & L8  & 0.591 & 0.037 & 407 & 6.74 \\
              & L10 & 0.600 & 0.037 & 444 & 2.57 \\
              & L16 & 0.610 & 0.043 & 372 & 5.42 \\
    \bottomrule
  \end{tabular}
\end{table}

\begin{table}[H]
\centering
\scriptsize
\setlength{\tabcolsep}{3.2pt}
\caption{Logistic-probe warnings under occlusion; abort metrics are descriptive only.}
  \label{tab:logistic_warnings}
  \begin{tabular}{@{}llcccc@{}}
    \toprule
    Policy & Layer & AUROC & AUPRC & Lead time & Onsets/ep \\
    \midrule
    Baseline & L8  & 0.952 & 0.229 & -- & -- \\
             & L10 & 0.801 & 0.092 & -- & -- \\
             & L16 & 0.972 & 0.352 & -- & -- \\
    \midrule
    No-op & L8  & 0.811 & 0.118 & 164 & 6.51 \\
          & L10 & 0.557 & 0.035 & 335 & 3.28 \\
          & L16 & 0.815 & 0.134 & 182 & 10.02 \\
    \midrule
    Abort & L8  & 0.722 & 0.148 & 0 & 0.88 \\
          & L10 & 0.645 & 0.142 & 0 & 0.77 \\
          & L16 & 0.731 & 0.170 & 0 & 0.91 \\
    \midrule
    Hold last & L8  & 0.809 & 0.103 & 168 & 6.18 \\
              & L10 & 0.552 & 0.034 & 349 & 3.40 \\
              & L16 & 0.740 & 0.091 & 197 & 8.44 \\
    \bottomrule
  \end{tabular}
\end{table}

\end{document}